\def\year{2017}\relax
\documentclass[letterpaper]{article}
\usepackage{aaai17}
\usepackage{times}
\usepackage{helvet}
\usepackage{courier}
\usepackage{graphicx} 
\usepackage{subfigure} 
\usepackage{algorithm}
\usepackage{algorithmic}
\usepackage{amsfonts}
\usepackage{amsmath}
\usepackage{commath}
\usepackage{array}
\begin{document}
%
\title{Generating Images Part by Part with Composite Generative Adversarial Networks}
\author{Hanock Kwak and Byoung-Tak Zhang \\ 
	School of Computer Science and Engineering  \\
	Seoul National University, Seoul 151-744, Korea \\
	\{hnkwak, btzhang\}@bi.snu.ac.kr}
\maketitle
\begin{abstract}
Image generation remains a fundamental problem of artificial intelligence, specifically in deep learning. The generative adversarial network (GAN) architecture was successful in generating high-quality samples of natural images. We propose a model called composite generative adversarial network (CGAN), that disentangles complicated factors of images with multiple generators in which each generator generates some part of the image. Those parts are combined by an alpha blending process to create a new single image. For example, it can generate background, face, and hair sequentially with three generators trained on face images. There is no supervision on what each generator should generate. The CGAN assigns roles for each generator by factorizing the common factors of images and creates realistic samples as good as GAN. Also, we combined a variational autoencoder with CGAN to visualize the sub-manifolds of latent space learned.
\end{abstract}

\noindent Images are composed of several different objects forming a hierarchical structure with various styles and shapes. Deep learning models are used to disentangle those complex underlying patterns \cite{reed2014learning}\cite{wang2016generative}, build distributed feature representations \cite{hinton2006reducing}, and solve classification \cite{krizhevsky2012imagenet} and generation \cite{radford2015unsupervised} problems using large datasets. While lots of classification tasks have focused on abstraction for a finite number of labels, generation tasks need to reconstruct raw images inversely from latent variables. To achieve such a task, the latent variables of generative models must contain detailed information about the raw images, which is not the case for the feature vectors in discriminative models. This gives us interesting challenges in generation task which is a fundamental problem of artificial intelligence. Even though we can easily imagine a scene by combining and mixing semantic parts, current generative models are far from reaching our abilities. 

Generative adversarial networks (GANs) \cite{goodfellow2014generative}, based on deep neural networks, are successful unsupervised learning models that can generate samples of natural images generalized from the training data. It provides an alternative to the intractable maximum likelihood estimation and pixel-wise loss functions. GAN simultaneously trains two models: a generator G that tries to generate real images, and a discriminator D that classifies between the real images that come from the training data and the fake images that come from G. The discriminator alleviates the lack of semantic consideration in pixel-wise loss function used in most auto-encoder models.

\begin{figure}[t]
	\vskip 0.2in
	\begin{center}
		\includegraphics[width=\columnwidth]{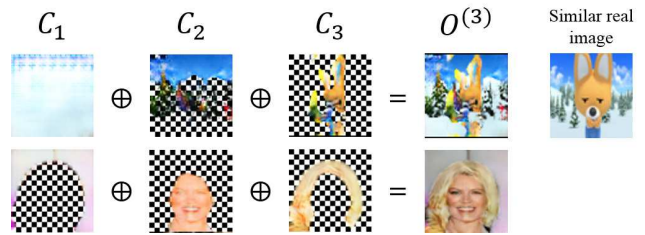}
		\caption{\textbf{Examples of generated images from CGAN.} $C_1, C_2, C_3$ are images generated from three generators, and $O^{(3)}$ is final output. Similar real images are shown for comparison. Black and white checkerboard is the default background for transparent images.}
		\label{mini_ex}
	\end{center}
	\vskip -0.1in
\end{figure} 

It is proven that if the GAN has enough capacity, data distribution formed by G converges to the distribution of real data \cite{goodfellow2014generative}. In practice, however, the convergence is intractable, and it is easy to overfit due to the exponential complexity of images in which multiple objects exist in any position with noisy features. To solve these issues, we propose a composite generative adversarial network (CGAN) that can generate images part by part instead of whole images directly. CGAN differs from other recurrent generative models \cite{mansimov2015generating,im2016generating}, which add the sequence of generated images intermixing each image in the overlapping areas. To address this problem, we used an alpha channel for opacity along with RGB channels to stack images iteratively with alpha blending process. The alpha blending process maintains the previous image in some areas and overlaps the new image entirely in other areas. For instance, given a transparent image the model may put a snowy background first, then later add trees and characters sequentially as shown in Figure \ref{mini_ex}.

Because the only inputs of CGAN are multiple latent variables from a prior distribution, it has its limitations while investigating the relation between latent space and the output images. We propose CGAN+VAE which is a combination of CGAN and variational autoencoder (VAE) \cite{kingma2013auto} as in \cite{DBLP:conf/icml/LarsenSLW16} to visualize such relation. We show that those latent variables form a sub-manifold conditioned on previous ones as described in Figure \ref{vae_mod}.

To illustrate how CGAN generates complicated images part by part, we used Oxford 102 Flowers, a dataset of flower images, and CelebA dataset consisting of face images. We also used a collection of cartoon videos for children, titled ``Pororo", where there are limited number of characters. 

Our contributions can be summarized as follows:

\begin{itemize}
	\item We propose a generative model, which generates images part by part and ends up with realistic images, for the first time.
	\item We visualized how multiple latent variables form a sub-manifold in CGAN without labels.
\end{itemize}

\begin{figure}[t]
	\vskip 0.2in
	\begin{center}
		\includegraphics[width=\columnwidth]{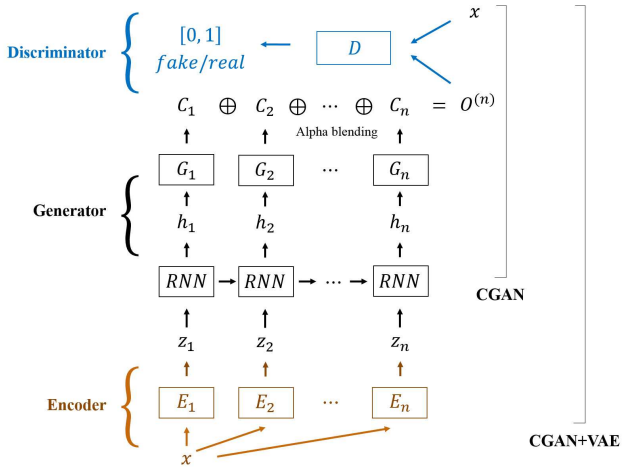}
		\caption{\textbf{The structure of CGAN and CGAN+VAE.} The discriminator and generator are main two components of CGAN. The RNN accepts a noise sequence $z_1, z_2, ..., z_n$ as input, and then recursively generates $h_1, h_2, ..., h_n$ which are then passed through each generator independently creating images $C_1, C_2, ..., C_n$ having RGBA channels. The images are then combined sequentially by the alpha blending process to form the final output $O^{(n)}$. Note that generators do not share weights, meaning that they are all different networks. We additionally combined variational autoencoder (VAE) to create $z_1, z_2, ..., z_n$ directly from the images.}
		\label{cgan_model}
	\end{center}
	\vskip -0.1in
\end{figure}

\section{Related Works}

Variants of probabilistic graphical models have been introduced to capture the underlying data distribution. Undirected graphical models with latent variables, such as restricted Boltzmann machines (RBMs) and deep Boltzmann machines (DBMs) \cite{salakhutdinov2009deep} provided the underlying discipline for pre-training deep neural networks. Deep belief networks (DBNs) \cite{hinton2006reducing} and its variants are hybrid models in which pre-trained DBMs and sigmoid belief networks are layer-wise mixed. DBNs can reproduce the input from multiple hidden layers, but they are restricted to a simple dataset due to the computationally costly step of Markov chain Monte Carlo (MCMC) methods. 

\cite{kingma2013auto} have proposed the variational auto-encoder (VAE) which has an encoder that approximates the posterior distribution of continuous latent variables, and a decoder that reconstructs the data from latent variables, trained by stochastic variational inference algorithm. \cite{DBLP:conf/icml/GregorDGRW15} has extended VAE to deep recurrent attention writer (DRAW) in which VAE is processed recurrently incorporating differentiable attention mechanism. Conditional alignDRAW model \cite{mansimov2015generating} is an extension of DRAW that generates an image conditioned on a sentence. \cite{gregor2016towards} also introduced a recurrent variational autoencoder architecture that significantly improves image modeling. These models differ from CGAN in that they construct images gradually from the first image through recurrent feedbacks. 

Similar to the DRAW, a recurrent adversarial network \cite{im2016generating} adds generated images from multiple generators sequentially and puts the sigmoid function at the end. Adding the images based on RGB channels results in intermixing of pixels. Our model uses the additional alpha channel to avoid this issue. 

Some of the variants of GAN, such as LAPGAN \cite{denton2015deep}, DCGAN \cite{radford2015unsupervised}, and recurrent adversarial network \cite{im2016generating}, improved the quality of generated images. VAE/GAN \cite{DBLP:conf/icml/LarsenSLW16} replaced pixel-wise loss function of VAE with feature-wise loss function where the features come from the discriminator of GAN. \cite{wang2016generative} has used two GAN: the Structure-GAN generates structures; the Style-GAN puts styles on the structures.

\section{Model} 

In this section, we first review GAN in detail and describe how GAN is extended with alpha blending process. We also designed CGAN+VAE model in which encoding capability is added to CGAN as in \cite{DBLP:conf/icml/LarsenSLW16}. 

In addition, several recent deep learning techniques (batch normalization \cite{conf/icml/IoffeS15}, ADAM \cite{kingma2014adam}, LSTM \cite{hochreiter1997long}, etc.), critical to the performance of CGAN, were utilized.

\subsection{Generative Adversarial Networks}

A GAN has two networks: a generator $G$ that tries to generate real data given noise $z \sim p_{z}(z)$, and a discriminator $D \in [0,1]$ that classifies the real data $x \sim p_{data}(x)$ and the fake data $G(z)$. The objective of $G$ is to fit the true data distribution deceiving $D$ by playing following minimax game:

\begin{equation}
\begin{split}
\min_{\theta_{G}} \max_{\theta_{D}} \quad & \mathbb{E}_{x \sim p_{data}(x)}[\log D(x)] \\
&+ \mathbb{E}_{z \sim p_{z}(z)}[\log(1 - D(G(z)))],
\end{split}
\end{equation}

where $\theta_{G}$ and $\theta_{D}$ are parameters of $G$ and $D$, respectively. Given a mini-batch of $\{x_1, x_2,...,x_m\}$ and $\{z_1, z_2,...,z_m\}$, $\theta_{G}$ and $\theta_{D}$ are updated for each iteration as following:

\begin{equation}
\begin{split}
& \mathcal{L}_{GAN} = \displaystyle\sum_{i=1}^{m} \log D(x_i) + \log(1 - D(G(z_i))),\\
& \theta_{G} \leftarrow \theta_{G} - \gamma_{G} \nabla_{\theta_{G}} \mathcal{L}_{GAN}, \\
& \theta_{D} \leftarrow \theta_{D} + \gamma_{D} \nabla_{\theta_{D}} \mathcal{L}_{GAN},
\end{split}
\end{equation}

where $\gamma_{G}$ and $\gamma_{D}$ are learning rates. We set different learning rates for $D$ and $G$ in practice to stabilize the learning progress.

\subsection{Alpha Blending}

The alpha blending combines two translucent images, producing a new blended image. The value of alpha is between $0.0$ and $1.0$, where a pixel is fully transparent if $0.0$, and fully opaque if $1.0$. We denote $C_{ij_{RGB}}$ as a 3-dimensional vector of RGB values in position $(i,j)$ and $C_{ij_{A}}$ as a scalar alpha value of the same position. CGAN uses alpha blending which covers the previous image $C^{(prev)}$ with the next image $C^{(next)}$ to make the new image $C^{(new)}$:

\begin{equation}
\begin{split}
C^{(new)}_{ij_{RGB}} = C^{(prev)}_{ij_{RGB}} C^{(prev)}_{ij_{A}} (1 - C^{(next)}_{ij_{A}}) + C^{(next)}_{ij_{RGB}} C^{(next)}_{ij_{A}}.
\end{split}
\end{equation}

Assuming that the new image is opaque, $C^{(new)}_{ij_{A}}$ is always $1$. This process maintains colors of the next image where $C^{(next)}_{ij_{A}}$ is nearly one, and that of the previous image where $C^{(next)}_{ij_{A}}$ is nearly zero.

\subsection{Composite Generative Adversarial Networks}

A CGAN, a extension of GAN, consists of multiple generators connected with a recurrent neural network (RNN) as shown in Figure \ref{cgan_model}. The generators in CGANs are different from that of GANs as there are additional alpha channels in the output. The images are then combined sequentially with alpha blending to form a final image. 

Given noise vectors $z_1, z_2, ..., z_n$ from predefined distribution, the RNN produces $h_1, h_2, ..., h_n$ to be used as input by each generator sequentially as following:

\begin{align}
&z_1, z_2, ..., z_n \sim p_z(z), \\
&h_t = RNN(h_{t-1}, z_t).
\end{align} 

The RNN preserves the consistency between the generators, so that the generated images are all related. We denote the generators and the generated images as $G_1, G_2, ..., G_n$, and $C^{(1)}, C^{(2)}, ..., C^{(n)}$, respectively. Note that $C^{(i)}$ is a RGBA image in which all pixels are four dimensional vectors. To illustrate how a final output image $O^{(n)}$ is made, we denote intermediate images as $O^{(1)}, O^{(2)}, ..., O^{(n-1)}$. The following explains how $O^{(n)}$ is formed:

\begin{align}
&C^{(t)} = G_i(h_t), \\
&O^{(t)}_{ij_{RGB}} = 
\begin{cases} 
C^{(1)}_{ij_{RGB}} C^{(1)}_{ij_{A}} \quad & \text{if} \quad t = 1 \\ 
O^{(t-1)}_{ij_{RGB}} (1 - C^{(t)}_{ij_{A}}) + C^{(t)}_{ij_{RGB}} C^{(t)}_{ij_{A}} \quad & \text{if} \quad t > 1 \\
\end{cases}
\end{align}

The objective of generators in whole is same as that of GAN, and the algorithm of CGAN is similar to that of GAN as illustrated in algorithm \ref{alg:cgan}.

\begin{algorithm}[tb]
	\caption{The algorithm of CGAN}
	\label{alg:cgan}
	\begin{algorithmic}
		\STATE {\bfseries Input:} dataset $X$, the number of generators $n$, mini-batch size $m$
		\STATE Initialize $\theta_{D}, \theta_{G_1}, \theta_{G_2}, ..., \theta_{G_n}$.
		\FOR{number of training iterations}
		\STATE Select $m$ data $\{x_1, x_2,...,x_m\} \subset X$ randomly
		\STATE Draw $m$ noises $\{z_1, z_2,...,z_m\} \sim p_z(z)$
		\STATE $\theta_{D} \leftarrow \theta_{D} + \gamma_{D} \nabla_{\theta_{D}} \mathcal{L}_{GAN}$
		\FOR{$i=1$ {\bfseries to} $n$}
		\STATE $\theta_{G} \leftarrow \theta_{G_i} - \gamma_{G} \nabla_{\theta_{G_i}}\mathcal{L}_{GAN}$
		\ENDFOR
		\ENDFOR
	\end{algorithmic}
\end{algorithm}

\begin{algorithm}[tb]
	\caption{The algorithm of CGAN+VAE}
	\label{alg:cganvae}
	\begin{algorithmic}
		\STATE {\bfseries Input:} dataset $X$, the number of generators $n$, mini-batch size $m$
		\STATE Initialize $\theta_{D},\theta_{E_1}, \theta_{E_2}, ..., \theta_{E_n}, \theta_{G_1}, \theta_{G_2}, ..., \theta_{G_n}$.
		\FOR{number of training iterations}
		\STATE Select $m$ data $\{x_1, x_2,...,x_m\} \subset X$ randomly
		\STATE Draw $m$ noises $\{z_1, z_2,...,z_m\} \sim p_z(z)$
		\STATE $\theta_{D} \leftarrow \theta_{D} + \gamma_{D} \nabla_{\theta_{D}} \mathcal{L}_{GAN}$
		\FOR{$i=1$ {\bfseries to} $n$}
		\STATE $\theta_{G_i} \leftarrow \theta_{G_i} - \nabla_{\theta_{G_i}}(\gamma_{G_{GAN}}\mathcal{L}_{GAN} + \gamma_{G_{VAE}}\mathcal{L}_{VAE})$
		\ENDFOR
		\FOR{$i=1$ {\bfseries to} $n$}
		\STATE $\theta_{E_i} \leftarrow \theta_{E_i} - \nabla_{\theta_{E_i}}(\gamma_{E}\mathcal{L}_{VAE})$
		\ENDFOR
		\ENDFOR
	\end{algorithmic}
\end{algorithm}

\begin{figure*}[t]
	\vskip 0.0in
	\begin{center}
		\includegraphics[width=\textwidth]{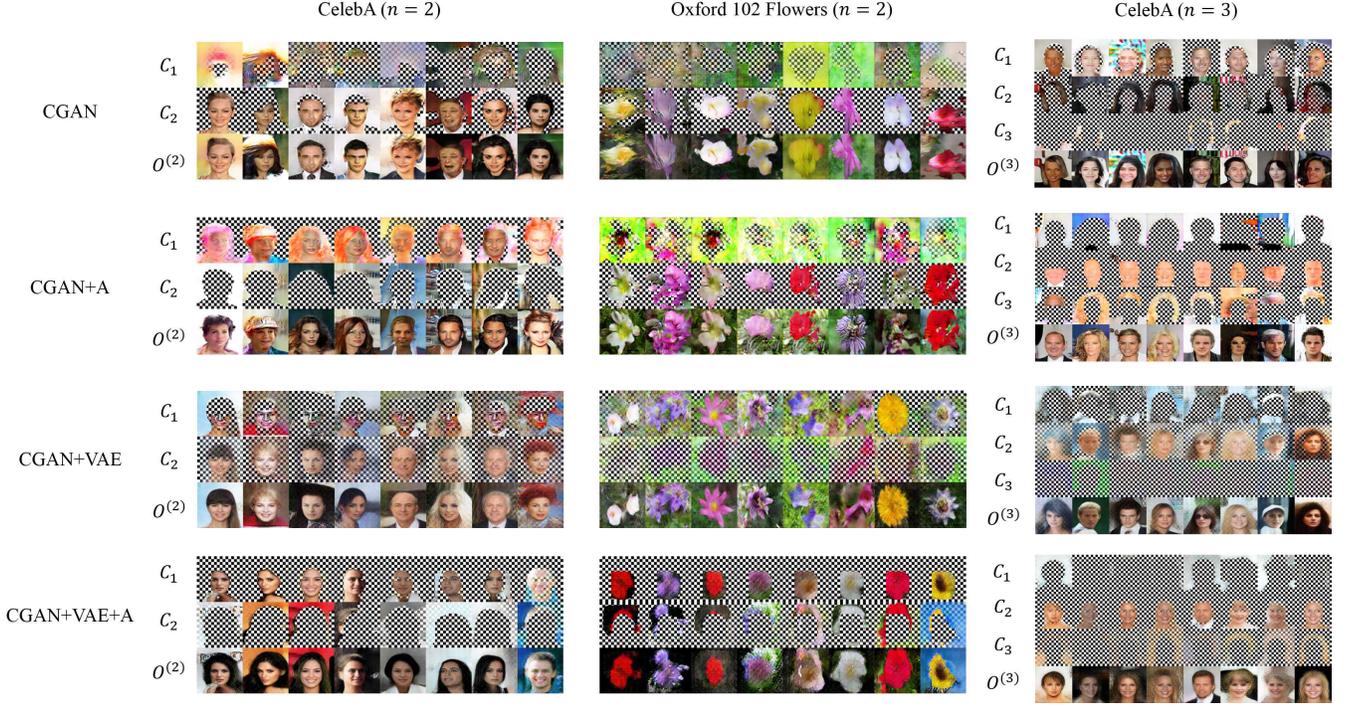}
		\caption{Samples from various CGANs after being trained on CelebA, and Oxfold 102 Flowers datasets. Black and white checkerboard is the default background for transparent images.}
		\label{samples_cgan}
	\end{center}
	\vskip 0.0in
\end{figure*} 

\subsection{Combining Variational Autoencoder}

A VAE \cite{kingma2013auto} is an autoencoder where the encoder is regularized by a prior over the latent distribution $p(z)$. The VAE is combined with CGAN adding $n$ encoders at bottom like Figure \ref{cgan_model}. Each encoder generates noise (or latent variable) $z_i$ and then remaining process is same as CGAN. Note that the prior $p(z)$ is same as the noise distribution of CGAN, so as to fit the latent space of two models.

The VAE maximizes log likelihood of data by maximizing a variational lower bound. The variational lower bound is calculated by subtracting prior regularization term from the reconstruction term:

\begin{align}
\log p(x_i) = -D_{KL}(q(z|x_i)||p(z)) + E_{q(z|x_i)}[\log p(x_i|z)].
\end{align}

The discriminator of GAN also can be used as a rich feature extractor. \cite{DBLP:conf/icml/LarsenSLW16} has shown that minimizing the reconstruction error expressed in a hidden layer of the discriminator improves the overall quality of the reconstructed images. Let $D_{h}(x)$ be a hidden vector of last convolutional layer of $D$. Then $p(D_{h}(x_i)|z)$ is defined as a normal distribution whose mean comes from the sample $\hat{x}$ of the decoder:

\begin{align}
 p(D_{h}(x_i)|z) = \mathcal{N}(D_{h}(x_i)|D_{h}(\hat{x}), I).
\end{align}

Putting it all together, given a mini-batch of $\{x_1, x_2,...,x_m\}$ we define VAE loss $\mathcal{L}_{VAE}$ as follows:

\begin{equation}
\begin{split}
\mathcal{L}_{VAE}  & = \displaystyle\sum_{i=1}^{m} D_{KL}(q(z|x_i)||p(z)) - E_{q(z|x_i)}[\log p(x_i|z)]\\
&  - E_{q(z|x_i)}[\log p(D_{h}(x_i)|z)] .
\end{split}
\end{equation}

\subsection{Alpha Loss}

Even though there are multiple generators, only a single generator can govern the final output image and fulfill the objective function. This problem frequently occurs in practice. To avoid this problem, we constrained sum of the alpha values to be same as a predefined bound $u$. Still all generators can generate similar images with small alpha values rather than heterogeneous images, so we also constrained each alpha value to be near zero or one: 

\begin{equation}
\begin{split}
\mathcal{L}_{\alpha} = \abs{u - \sum_{i, j}C_{ij_{A}}} + \sum_{i, j}-(C_{ij_{A}} - 0.5)^2 .
\end{split}
\end{equation}

The models using alpha loss are named as CGAN+VAE+A and CGAN+A.

\section{Experiments}

All images are resized to $64\times 64$ with antialiasing. We used long short-term memory (LSTM) \cite{hochreiter1997long} architecture for RNN. The structures of each generator and discriminator are similar to that of DCGAN \cite{radford2015unsupervised} which has series of four fractionally strided convolutions (or transposed convolution). We chose a multivariate normal distribution as the prior $p_z(z)$.

\subsection{Dataset}
\subsubsection{CelebA face images}

CelebA \cite{liu2015faceattributes} dataset contains 202,599 face images and 11,177 number of identities. The images in this dataset cover large pose variations and background clutter. 

\subsubsection{Oxford 102 Flowers}

Oxford 102 Flowers \cite{Nilsback08} is a dataset of 102 kinds of flowers commonly found in the United Kingdom. Each class consists of between 40 and 258 images. The images have large scale, pose and light variations.

\subsubsection{Pororo cartoon video}

Pororo is a cartoon video with 1,232 minutes total running time. The dataset has large diversities of poses, sizes, and positions of characters, yet most of the backgrounds are snowy mountains, glaciers, forests and wooden houses. We captured frames for each second of the video and shuffled to avoid bias. 

\subsection{Evaluation}
Evaluation of generative models is problematic \cite{Theis2016a} due to various objectives (density estimator, feature learning, clustering, etc.) of unsupervised learning. Since our target is to generate images part by part, qualitative analysis takes the most part in the assessment of CGAN. 

To measure quality of the images, a Structured Similarity Index (SSIM) \cite{wang2004image} is used as a quantitative measure rather than the pixel-wise error. SSIM is a perception-based model that incorporates important perceptual phenomena, including both luminance masking and contrast masking terms. It is measured by taking small windows in images to compare the structural information of the two images locally. To evaluate the GAN based models, we compared samples to the test data with SSIM. Only the largest SSIM value among the samples is taken into account for each test data:

\begin{equation}
\begin{split}
Q = \frac{1}{N}\sum_{i=1}^{N}\max_{s\in S}SSIM(s, x_i),
\end{split}
\end{equation}

where $S$ is a set of samples and $x_1, x_2, ..., x_N$ are test data. SSIM is ranged from $-1$ to $1$. Larger values mean that the two images are more similar.

Table \ref{ssi_table} shows that quality of the output images from CGANs are similar to GAN. 

\begin{table*}[t]
	\centering
	\caption{SSIM measure on various datasets and the number of generators.}
	\label{ssi_table}
	\setlength\extrarowheight{3pt}
	\begin{tabular}{cccccc}
		\hline
		\multicolumn{1}{c|}{\textbf{}}   & \multicolumn{5}{c}{\textbf{Dataset and the \# of G}}   \\ \hline
		\multicolumn{1}{c|}{\textbf{Model}}   & \textbf{CelebA (n=2)} & \textbf{CelebA (n=3)} & \textbf{Pororo (n=2)} & \textbf{Pororo (n=3)} & \textbf{102 Flowers (n=2)}  \\ \hline  \hline
		\multicolumn{1}{c|}{\textbf{GAN}}    & \multicolumn{2}{c}{$0.449_{\pm0.077}$} & \multicolumn{2}{c}{$0.406_{\pm0.15}$}   & $0.296_{\pm0.069}$   \\ \hline
		\multicolumn{1}{c|}{\textbf{CGAN}}       & $0.443_{\pm0.075}$  & $0.443_{\pm0.077}$  & $0.391_{\pm0.14}$  & $0.396_{\pm0.15}$  & $0.290_{\pm0.069}$     \\ 
		\multicolumn{1}{c|}{\textbf{CGAN+A}}     & $0.437_{\pm0.073}$  & $0.448_{\pm0.077}$  & $0.375_{\pm0.13}$  & $0.393_{\pm0.14}$  & $0.265_{\pm0.069}$     \\
		\multicolumn{1}{c|}{\textbf{CGAN+VAE}}   & $0.444_{\pm0.075}$  & $0.443_{\pm0.076}$  & $0.396_{\pm0.15}$  & $0.393_{\pm0.15}$  & $0.276_{\pm0.061}$     \\
		\multicolumn{1}{c|}{\textbf{CGAN+VAE+A}} & $0.449_{\pm0.075}$  & $0.443_{\pm0.075}$  & $0.393_{\pm0.13}$  & $0.386_{\pm0.14}$  & $0.286_{\pm0.068}$    \\ \hline
	\end{tabular}
\end{table*}

\begin{figure}[t]
	\vskip 0.2in
	\begin{center}
		\includegraphics[width=\columnwidth]{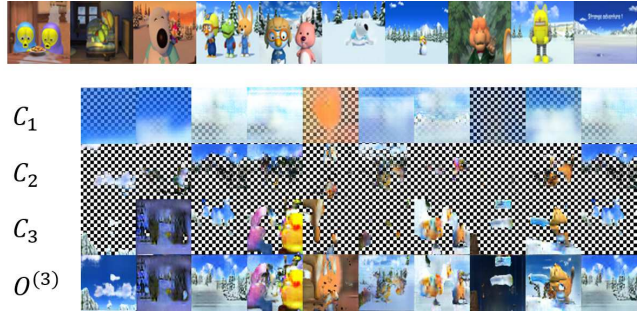}
		\caption{\textbf{Samples generated from CGAN after being trained on Pororo dataset.} The top row shows real images and the below four rows show how CGAN generates images. Note that some characters are generated in $C_3$.}
		\label{samples_pororo}
	\end{center}
	\vskip -0.1in
\end{figure}

\subsection{Generating Images Part by Part}
We used up to three generators, and the result is shown in Figure \ref{samples_cgan}.  Even though the overall process is stochastic and behavior of CGAN is unpredictable due to its unsupervised setting, CGAN successfully generates images with all generators sharing their efforts impartially. In samples from CGAN+A for CelebA $(n=3)$, $G_1, G_2, G_3$ generated backgrounds, faces, and hair parts, respectively, to end up with the final images. In the case of flowers, two generators have shared the task by generating backgrounds and flowers separately. CGAN also works for complicated cartoon videos such as Pororo. The results are shown in Figure \ref{samples_pororo}. 

In the samples from CGAN for CelebA $(n=3)$, the third generator failed to generate meaningful images. Applying alpha loss on that CGAN, the problem diminishes. Also in the case of CGAN+VAE, using alpha loss forces intermediate images to be less blurry and more separable.

\subsection{Disentanglement of Factors in Images}

\begin{figure}[t]
	\vskip 0.2in
	\begin{center}
		\includegraphics[width=\columnwidth]{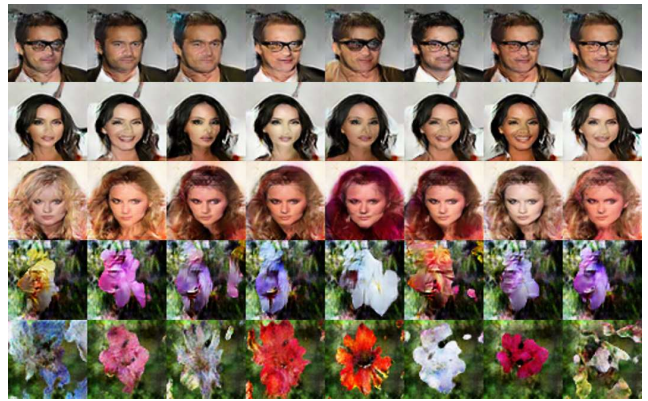}
		\caption{\textbf{Samples generated from fixed $z_1$ and random $z_2, z_3, ..., z_n$.} The faces are generated from CGAN+A $(n = 3)$ and the flowers are generated from CGAN $(n = 2)$. Each row shares same $z_1$.}
		\label{fix_z}
	\end{center}
	\vskip -0.1in
\end{figure}

\begin{figure}[t]
	\vskip 0.2in
	\begin{center}
		\includegraphics[width=\columnwidth]{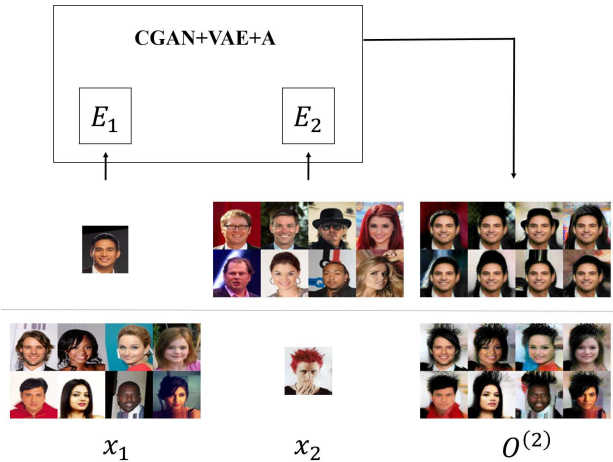}
		\caption{\textbf{The relation between latent variables and output images in CGAN+VAE+A.} Various images are encoded through one encoder while fixing the image for the other encoder. The first images $x_1$ are reconstructed as $O^{(2)}$ reflecting the features of the second images $x_2$. }
		\label{vae_mod}
	\end{center}
	\vskip -0.1in
\end{figure}

The latent variables $z_1, z_2, ..., z_n$ of CGAN represent the factors of images. Since they are passed through a sequential model RNN, the input $h_n$ is dependent to previous $z_1, z_2, ..., z_{n-1}$. The first latent variable $z_1$ determines overall outline of the images and the other latent variables manipulate rest of the variations conditioned on $z_1$ as seen in Figure \ref{fix_z}. To visualize the learned space more precisely, VAE was utilized to create latent variables from images. Figure \ref{vae_mod} shows how common factors influenced the final images. CGAN successfully disentangles the factors of images with no labels.

\section{Discussion}

Disentanglement of high dimensional data has been thoroughly studied, yet it has not been sufficiently solved using unsupervised learning. Compared to humans who see the video of real life for numerous years, 100K $\sim$ 1M of images might be too small for such learning tasks. We found that it is possible to learn interacting factors from images without any labels by constructing the hierarchical structures of the images. With a larger dataset, future models could understand images in more detail and might be capable of changing the arrangement of objects dynamically without touching the backgrounds. 

Our model can be extended to other domains such as video, text, audio, or a combination of them using specialized encoders and generators. One of the primary goals of unsupervised learning is any modality to any modality mapping. Since most data has hierarchical structures, studies on decomposing the combined data are essential.

\bibliographystyle{named}
\bibliography{aaai17}

\end{document}